\documentclass[twocolumn]{svjour3}          % twocolumn
\smartqed  % flush right qed marks, e.g. at end of proof
\usepackage[british]{babel}

\usepackage{graphicx}

\usepackage{natbib}
\setcitestyle{numbers,square,comma}

\usepackage{hyperref}
\usepackage{todonotes}
\usepackage{verbatim}

\usepackage{caption}
\usepackage{subcaption}

\usepackage{balance}

\usepackage{enumitem}
\usepackage{hyphenat}

\begin{document}

\title{One Explanation Does Not Fit All}
\subtitle{The Promise of Interactive Explanations for Machine Learning Transparency}

\author{Kacper Sokol%
\and%
Peter Flach%
}

\institute{Kacper Sokol \at
              Department of Computer Science, University of Bristol\\
              Bristol, United Kingdom\\
              \email{K.Sokol@bristol.ac.uk}
           \and
           Peter Flach \at
              Department of Computer Science, University of Bristol\\
              Bristol, United Kingdom\\
              \email{Peter.Flach@bristol.ac.uk}
}

%\journalname{test}
%\date{Received: date / Accepted: date}
\date{(Published in the \emph{K\"unstliche Intelligenz} journal, special issue on \emph{Challenges in Interactive Machine Learning}.)}
% The correct dates will be entered by the editor
%
%
\maketitle

\sloppy

\begin{abstract}
The need for transparency of predictive systems based on Machine Learning algorithms arises as a consequence of their ever\hyp{}increasing proliferation in the industry. %
Whenever black\hyp{}box algorithmic predictions influence human affairs, the inner workings of these algorithms should be scrutinised and their decisions explained to the relevant stakeholders, including the system engineers, the system's operators and the individuals whose case is being decided. %
While a variety of interpretability and explainability methods is available, none of them is a panacea that can satisfy all diverse expectations and competing objectives that might be required by the parties involved. %
We address this challenge in this paper by discussing the promises of \emph{Interactive} Machine Learning for improved transparency of black\hyp{}box systems using the example of contrastive explanations -- a state\hyp{}of\hyp{}the\hyp{}art approach to \emph{Interpretable} Machine Learning.%

Specifically, we show how to personalise counterfactual explanations by interactively adjusting their conditional statements and extract additional explanations by asking follow\hyp{}up ``What if?'' questions. %
Our experience in building, deploying and presenting this type of system allowed us to list desired properties as well as potential limitations, which can be used to guide the development of interactive explainers. %
While customising the medium of interaction, i.e., the user interface comprising of various communication channels, may give an impression of personalisation, we argue that adjusting the explanation itself and its content is more important. %
To this end, properties such as breadth, scope, context, purpose and target of the explanation have to be considered, in addition to explicitly informing the explainee about its limitations and caveats. %
Furthermore, we discuss the challenges of mirroring the explainee's mental model, which is the main building block of intelligible human\hyp{}machine interactions. %
We also deliberate on the risks of allowing the explainee to freely manipulate the explanations and thereby extracting information about the underlying predictive model, which might be leveraged by malicious actors to steal or game the model. %
Finally, building an end\hyp{}to\hyp{}end interactive explainability system is a challenging engineering task; unless the main goal is its deployment, we recommend ``Wizard of Oz'' studies as a proxy for testing and evaluating standalone interactive explainability algorithms.%
% \PACS{PACS code1 \and PACS code2 \and more}
% \subclass{MSC code1 \and MSC code2 \and more}
\end{abstract}

\section{Introduction\label{sec:chap07:intro}}
Given the opaque, ``black\hyp{}box'' nature of complex Machine Learning (ML) systems, their deployment in mission\hyp{}critical domains is limited by the extent to which they can be interpreted or validated. %
In particular, predictions, (trained) models and (training) data should be accounted for. %
One way to achieve this is by ``transparency by design'', so that all components of a predictive system are ``glass boxes'', i.e., ante\hyp{}hoc transparency~\cite{rudin2019stop}. %
Alternatively, transparency might be achieved with post\hyp{}hoc tools, which have the advantage of not limiting the choice of a predictive model in advance~\cite{ribeiro2016why}. %
The latter approaches can either be model\hyp{}specific or model\hyp{}agnostic~\cite{robnik2018perturbation}. %
Despite this wide range of available tools and techniques, many of them are non\hyp{}interactive, providing the explainee (a recipient of an explanation) with a single explanation that has been optimised according to some predefined metric. %
While some of these methods simply cannot be customised by the end user without an in\hyp{}depth understanding of their inner workings, others can take direct input from users with a varying level of domain expertise: from a lay audience -- e.g., selecting regions of an image in order to query their influence on the classification outcome -- to domain experts -- e.g., tuning explanation parameters such as the importance of neighbouring data points. %
A particular risk of a lack of interaction and personalisation mechanisms is that the explanation may not always align with users' expectations, reducing its overall value and usefulness.%

Allowing the user to guide and customise an explanation can benefit the transparency of a predictive system by making it more suitable and appealing to the explainee, for example, by adjusting its content and complexity. %
Therefore, personalisation can be understood as modifying an explanation or an explanatory process to answer user\hyp{}specific questions. %
For counterfactual explanations of the form: ``had feature \(X\) been different, the prediction of the model would have been different too'', these can be user\hyp{}defined constrains on the number and type of features (\(X\)) that can and cannot appear in the conditional statement. %
Delegating the task of customising and personalising explanations to the end user via interaction mitigates the need for the difficult process of modelling the user's mental model beforehand, rendering the task feasible and making the whole process feel more natural, engaging and less frustrating.%

In human interactions, understanding is naturally achieved via an \emph{explanatory dialogue}~\cite{miller2018explanation}, possibly supported with visual aids. %
Mirroring this explanatory process for ML transparency would make it attractive and accessible to a wide audience. %
Furthermore, allowing the user to customise explanations extends their utility beyond ML transparency. %
The explainee can steer the explanatory process to inspect fairness (e.g., identify biases towards protected groups\footnote{A \emph{protected group} is a sub\hyp{}population in a data set created by fixing a value of a protected attribute such as \emph{age}, \emph{gender} or \emph{ethnicity}, which discriminating upon is illegal.})~\cite{kusner2017counterfactual}, assess accountability (e.g., identify model errors such as non\hyp{}monotonic predictions with respect to monotonic features)~\cite{lipton2017doctor} or debug predictive models~\cite{kulesza2015principles,sokol2019counterfactual}. %
In contrast to ML tasks~\cite{kapoor2010interactive} -- where any interaction may be impeded by human\hyp{}incomprehensible internal representations utilised by a predictive model -- interacting with explainability systems is feasible as the representation has to be human\hyp{}understandable in the first place, thereby enabling a bi\hyp{}directional \emph{communication}. %
Interaction with explanatory systems also allows incorporating new knowledge into the underlying ML algorithm and building a mental model of the explainee, which will help to customise the resulting explanations.%

\begin{figure*}[htb]
    \centering
    \begin{subfigure}[b]{0.45\textwidth}
        \centering
        \includegraphics[width=\textwidth]{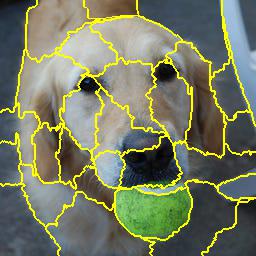}
        \caption{Default segmentation.}
    \end{subfigure}
    \hfill
    \begin{subfigure}[b]{0.45\textwidth}
        \centering
        \includegraphics[width=\textwidth]{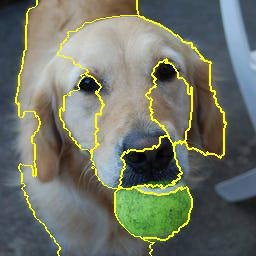}
        \caption{User\hyp{}merged segmentation.}
    \end{subfigure}
    
    \caption{Surrogate explainers of image classifiers require an interpretable representation -- image segmentation -- to communicate the explanation to the user. These explainers try to identify segments of an image that influence its classification the most, i.e., segments of high importance. Since the default outcome of image segmentation (a) may be unintuitive, we encourage the explainee to personalise the segmentation (b) to represent meaningful concepts.\label{fig:chap07:segmentation}}
\end{figure*}

Consider the example of explaining an image with a local surrogate method that relies upon super\hyp{}pixel segmentation (e.g., LIME algorithm introduced by \citet{ribeiro2016why}). %
While super\hyp{}pixel discovery may be good at separating colour patches based on their edges, these segments do not often correspond to meaningful \emph{concepts} such as ears or a tail for a dog image -- see Figure~\ref{fig:chap07:segmentation} for an example. %
The explanation can be personalised by allowing the explainee to \emph{merge} and \emph{split} segments before analysing their influence on the output of a black\hyp{}box model, thereby implicitly answering what prompted the explainee to alter the segmentation. %
User input is a welcome addition given the complexity of images; a similar approach is possible for tabular and text data, although user input is often unnecessary in these two cases. %
For tabular data the explainee may select certain feature values that are of interest or create meaningful binning for some of the continuous features; for text data (treated as a bag of words) the user may group some words into a phrase that conveys the correct meaning in that particular sentence. %
This exchange of knowledge between the explainee and the explainability system can considerably increase the quality of explanations, but also poses a significant safety, security and privacy risk. %
A malicious explainee may use such a system to reveal sensitive data used to train the underlying predictive (or explanatory) model, extract proprietary model components, or learn its behaviour in an attempt to game it (see Section~\ref{sec:chap07:glass-box_desiderata}).%

After \citeauthor{miller2018explanation}'s~\cite{miller2018explanation} seminal work -- inspired by explanation research in the social sciences -- drew attention to the lack of human\hyp{}aspect considerations in the eXplainable Artificial Intelligence (XAI) literature -- with many such systems being designed by the technical community for the technical community~\cite{miller2017explainable} -- researchers started acknowledging the end user when designing XAI solutions. %
While this has advanced human\hyp{}centred design and validation of explanations produced by XAI systems, another of \citeauthor{miller2018explanation}'s insights received relatively little attention: the interactive, dialogue\hyp{}like nature of explanations. %
Many of the state\hyp{}of\hyp{}the\hyp{}art explainability approaches are static, one\hyp{}off systems that do not take user input or preferences into consideration beyond the initial configuration and parametrisation~\cite{friedman2001greedy,goldstein2015peeking,lundberg2017unified,ribeiro2016why,ribeiro2018anchors,wachter2017counterfactual}.\footnote{To clarify, the notion of interaction is with respect to the explanation, e.g., the ability of the explainee to personalise it, and not the overall interactiveness of the explainability system.} %
While sometimes the underlying explanatory algorithms are simply incapable of a meaningful interaction, others do apply a technique or utilise an explanatory artefact that can support it in principle. %
Part of this trend can be attributed to the lack of a well\hyp{}defined protocol for evaluating interactive explanations and the challenging process of assessing their quality and effectiveness, which -- in contrast to a one\hyp{}shot evaluation -- is a software system engineering challenge\footnote{%
Building such systems requires a range of diverse components: user interface, natural language processing unit, natural language generation module, conversation management system and a suitable and well\hyp{}designed XAI algorithm. %
Furthermore, most of these components are domain\hyp{}specific and cannot be generalised beyond the selected data set and use case.%
} and requires time\hyp{} and resource\hyp{}consuming user studies.%

\citet{schneider2019personalized} noted that bespoke explanations in AI -- achieved through interaction or otherwise -- are largely absent within the existing literature. %
Research in this space usually touches upon three aspects of ``personalised'' explanations. %
First, there are interactive machine learning systems where the user input is harnessed to improve performance of a predictive model or align the data processing with its operator's prior beliefs. %
While the classic active learning paradigm dominates this space, \citet{kulesza2015principles} designed a system that presents its users with classification explanations to help them refine and personalise the predictive task, hence focusing the interaction on the underlying ML model and not the explanations. %
Similarly, \citet{kim2015ibcm} introduced an interactive ML system with an explainability component, allowing its users to alter the data clustering based on their preferences. %
Secondly, the work of \citet{krause2016interacting} and \citet{weld2018challenge} focused on interactive (multi\hyp{}modal) explainability systems. %
Here, the interaction allows the explainee to elicit more information about an ML system by receiving a range of diverse explanations derived from a collection of XAI algorithms such as Partial Dependence (PD)~\cite{friedman2001greedy} and Individual Conditional Expectation (ICE)~\cite{goldstein2015peeking} plots. %
While this body of research illustrates what such an interaction (with multiple explanatory modalities) might look like and persuasively argues its benefits~\cite{weld2018challenge}, the advocated interaction is mostly with respect to the presentation medium itself -- e.g., an interactive PD plot -- and cannot be used to customise and personalise the explanation \emph{per se}. %
Thirdly, \citet{madumal2019grounded} and \citet{schneider2019personalized} developed theoretical frameworks for interactive, personalised explainability that prescribe the interaction protocol and design of such systems. %
However, these theoretical foundations have not yet been utilised to design and implement an interactive explainability system coherent with XAI desiderata outlined by \citet{miller2018explanation}, which could offer customisable explanations. %
A more detailed overview and discussion of the relevant literature is given in Section~\ref{sec:chap07:related_work}.%

In this paper we propose an architecture of a truly interactive explainability system, demonstrate how to build such a system, analyse its desiderata, and examine how a diverse range of explanations can be personalised (Section~\ref{sec:chap07:glass-box}). %
Furthermore, we discuss lessons learnt from presenting it to both a technical and a lay audience, and provide a plan for future research in this direction (Section~\ref{sec:chap07:discussion}). %
As a first attempt to build an XAI system that allows the explainee to customise and personalise the explanations, we decided to use a \emph{decision tree} as the underlying predictive model. %
This choice simplifies many steps of our initial study, allowing us to validate (and guarantee correctness of) the explanations and reduce the overall complexity of the explanation generation and tuning process by inspecting the structure of the underlying decision tree. %
Using \emph{ante\hyp{}hoc} explanations derived from a single predictive model also allows us to mitigate engineering challenges that come with combining multiple independent XAI algorithms as proposed by \citet{weld2018challenge}. %
Furthermore, a decision tree can provide a wide range of diverse explanation types, many of which can be customised and personalised. %
Specifically, for global model explanations we provide%
\begin{itemize}%
    \item \emph{model visualisation}, and%
    \item \emph{feature importance};%
\end{itemize}%
while as prediction explanations we provide%
\begin{itemize}%
  \item \emph{decision rule} -- extracted from a root\hyp{}to\hyp{}leaf path,%
  \item \emph{counterfactual} -- achieved by comparing decision rules for different tree leaves, and%
  \item \emph{exemplar} -- a similar training data point extracted from the tree leaves.%
\end{itemize}%

When presented to the user, all of these explanations span a wide range of explanatory artefacts in visualisation (images) and textualisation (natural language) domains, thereby allowing us to test the extent to which they can be interactively personalised. %
Contrastive explanations, in particular \emph{class\hyp{}contrastive} counterfactual statements, are the foundation of our system. %
These take the form of: ``had \emph{one of the attributes been different in a particular way}, the classification outcome would have changed as follows\ldots.'' %
Arguably, they are the most suitable, natural and appealing explanations targeted at humans~\cite{miller2018explanation,wachter2017counterfactual}. %
In addition to all of their desired properties grounded in the social sciences~\cite{miller2018explanation} and legal considerations~\cite{wachter2017counterfactual}, they can be easily adapted to an interactive dialogue aimed at personalisation, which is not widely utilised. %
In our system they are delivered in an interactive dialogue -- a natural language conversation, which is the most intuitive explanatory mechanism~\cite{miller2018explanation}. %
In summary, our approach aims to build a holistic and diverse interactive XAI system where the interaction is focused on \textbf{personalising} explanations (in accordance with \citeauthor{miller2018explanation}'s~\cite{miller2018explanation} notion of XAI interactivity) as opposed to simply building an XAI system that provides explanations interactively (to explain different aspects of a black\hyp{}box system using a range of XAI algorithms) -- a subtle but significant difference.%

\section{Background and Related Work\label{sec:chap07:related_work}}
Throughout our research we have identified three distinct research strands in the literature that are relevant to interactive explanations:%
\begin{itemize}
    \item interactive Artificial Intelligence and Machine Learning (mostly from the perspective of Human\hyp{}Computer Interaction),%
    \item interactive explainability tools, which are interactive with respect to the user interface that delivers the explanations, and%
    \item theory of explanatory interactions, e.g., a natural language dialogue, between two intelligent agents (be them humans, machines or one of each).%
\end{itemize}

The Human\hyp{}Computer Interaction community has identified the benefits of human input for tools powered by AI and ML algorithms that extend beyond the active learning paradigm where people act as data labelling oracles~\cite{amershi2014power}. %
For example, consider a movie recommendation system where the user provides both explicit feedback, such as movie ratings, and implicit feedback, e.g., movies that the user did not finish watching. %
In order to utilise the full potential of any feedback and ensure user satisfaction, the users have to understand how their input affects the system (in particular, its underlying predictive model). %
Among others, the users should be informed whether their feedback is incorporated into recommendations immediately or with a delay and how does ``liking'' a movie influences future recommendations (e.g., similar genre and shared cast members). %
Here, this understanding is mostly achieved (in the case of user studies) by inviting the users to onboarding sessions or (progressively) disclosing relevant information via the user interface, hence the explanation is provided outside of the autonomous system. %
These actions help the users build a correct mental model of the ``intelligent agent'' allowing them to seamlessly interact with it. %
Ideally, the users would develop a \emph{structural} mental model that gives them a deep and in\hyp{}detailed understanding of how the ML or AI operates, however a \emph{functional} mental model (a shallow understanding) often suffices.%

While explanations are often provided outside of the interactive agents, several researchers showed how to integrate them into the user interface of autonomous systems~\cite{kulesza2010explanatory,kulesza2013too,kulesza2015principles,kim2015ibcm}. %
This is especially useful when the system is dynamic -- e.g., its underlying predictive model evolves over time -- in which case the explanations support and inform users' interaction with the system and guide the users towards achieving the desired objective. %
There are two prominent examples of such systems in the literature. %
\citet{kulesza2015principles} developed an interactive topic\hyp{}based Na\"ive Bayes classifier for electronic mail to help the users ``debug'' and ``personalise'' the categorisation of emails. %
The users are presented with explanations pertaining to every classified email -- words in the email that contribute towards and against a given class -- and are allowed to adjust the weights of these factors if they do not agree with their premise, and hence refine and personalise the model in a process which the authors call \emph{explanatory debugging}~\cite{kulesza2010explanatory,kulesza2013too,kulesza2015principles}. %
\citet{kim2015ibcm} designed a similar system where the users can interactively personalise clustering results -- which are explained with cluster centroids and prominent exemplars -- by promoting and demoting data points within each cluster. %
In this literature, explanations of predictive models are used to improve users' understanding (mental model) of an autonomous system to empower them to better utilise its capabilities (e.g., via improved personalisation) by interactively providing beneficial input. %
Hence, AI and ML explainability is not the main research objective in this setting and the explanations are not interactive themselves.%

The second research strand that we identified in the literature covers interactive, multi\hyp{}modal explainability tools in AI and ML. %
These systems allow investigating a black\hyp{}box model and its predictions by providing the user with a variety of explanations produced with a range of diverse explainability techniques delivered via (an interactive) user interface. %
For example, \citet{krause2016interacting} built an interactive system that allows its users to inspect Partial Dependence~\cite{friedman2001greedy} of selected features (model explanation) and investigate how changing feature values for an individual data point would affect its classification (prediction explanation)~\cite{krause2016interacting,krause2016using}. %
Whereas combining multiple explainability techniques within a single system with a unified user interface is feasible, ensuring coherence of the diverse explanations that they produce poses significant challenges as some of the explanations may be at odds with each other and provide contradictory evidence for the same outcome. %
\citet{weld2018challenge} showed an idealised example of such a system and persuasively argued its benefits, however they have not discussed how to mitigate the issue with contradictory and competing explanations. %
While both of these explainability tools are \emph{interactive}, the interaction is limited to the presentation medium of the explanations and a choice of explainability technique, which, we argue, is insufficient -- the system is interactive but the explanation is not. %
\emph{Truly interactive} explanations allow the user to tweak, tune and personalise them (i.e., their content) via an interaction, hence the explainee is given an opportunity to guide them in a direction that helps to answer selected questions.%

The third research strand in the literature characterises explanatory communication as interaction between two intelligent agents~\cite{arioua2015formalizing,walton2016dialogue,madumal2019grounded,schneider2019personalized}. %
\citet{arioua2015formalizing} formalised explanatory dialogues in Dung's argumentation framework~\cite{dung2009assumption} and introduced ``questioning'' dialogues to evaluate success of explanations. %
\citet{walton2016dialogue} introduced a similar \emph{shift model} composed of two distinct dialogue modes: an explanation dialogue and an examination dialogue, where the latter is used to evaluate the success of the former~\cite{walton2007dialogical,walton2011dialogue,walton2016dialogue}. %
\citet{madumal2019grounded} refined these two approaches and proposed an interactive communication schema that supports explanatory and questioning dialogues, which also allow the explainee to formally challenge and argue against some of the decisions and their explanations. %
\citet{madumal2019grounded} have also empirically evaluated their explanatory dialogue protocol on text corpora to show its flexibility and applicability to a range of different scenarios. %
\citet{schneider2019personalized} approached this problem on a more conceptual level discussing interactions with various explainability tools and showing examples of how they could allow for personalised explanation. %
Most of the work presented in this body of literature is purely theoretical and has not yet been embraced by practical explainability tools.%

These diverse research strands come together to help eXplainable AI and Interpretable Machine Learning (IML) researchers and practitioners design appealing and useful explainability tools with many of their recommendations originating from explanatory interactions between humans. %
\citet{miller2018explanation} reviewed a diverse body of social sciences literature on human explanations and proposed an agenda for human\hyp{}centred explanations in AI and ML. %
\citet{miller2017explainable} noticed that explainability systems built for autonomous agents and predictive systems rarely ever consider the end users and their expectations as they are mostly ``built by engineers, for engineers.'' %
Since then, XAI and IML research has taken a more human\hyp{}centred direction, with many academics and engineers~\cite{wachter2017counterfactual,waa2018contrastive,schneider2019personalized,weld2018challenge,henin2019towards} evaluating their approaches against \citeauthor{miller2018explanation}'s guidelines to help mitigate such issues.%

Two of \citeauthor{miller2018explanation}'s recommendations are of particular importance: interactive, dialogue\hyp{}like nature of explanations and popularity of contrastive explanations among humans. %
While interactivity of explanations~\cite{schneider2019personalized} has been investigated from various viewpoints in the literature (and discussed earlier in this section), explanations delivered in a bi\hyp{}directional conversation, giving the explainee the opportunity to customise and personalise them, have not seen much uptake in practice. %
One\hyp{}off explanations are still the most popular operationalisation of explainability algorithms~\cite{schneider2019personalized}, where the explainer outputs a one\hyp{}size\hyp{}fits\hyp{}all explanation in an attempt to make the behaviour of a predictive system transparent. %
A slight improvement over this scenario is to enable the explainer to account for user preferences when generating the explanations~\cite{lakkaraju2019faithful,poyiadzi2019face}, but this modality is not common either. %
Interactively personalising an explanation allows the users to adjust its complexity to suit their background knowledge, experience and mental capacity; for example, explaining a disease to a medical student should take a very different form from explaining it to a patient. %
Therefore, an interactive system can satisfy a wide range of explainees' expectations, including objectives other than improving transparency such as inspecting individual fairness of algorithmic predictions~\cite{kusner2017counterfactual}.%

The prominence of contrastive statements in human explanations is another important insight from the social sciences, which also highlights their capacity to be interactively customised and personalised. %
In the recent years this type of explanations has proliferated into the XAI and IML literature in the form of \emph{class contrastive counterfactual} explanations: ``had you earned twice as much, your loan application would have been successful.'' %
This uptake can also be attributed to their legal compliance~\cite{wachter2017counterfactual} with the ``right to explanation'' introduced by the European Union's General Data Protection Regulation (GDPR). %
However, their capacity to be customised and personalised is often overlooked in practice~\cite{miller2018explanation,wachter2017counterfactual,waa2018contrastive,poyiadzi2019face}.%

All in all, many of \citeauthor{miller2018explanation}'s~\cite{miller2018explanation} insights from the social sciences have found their way into research and practice. %
An example of the latter is Google's \emph{People + AI Guidebook}\footnote{\url{https://pair.withgoogle.com}} describing best practices for designing human\hyp{}centred AI and ML products and acknowledging the importance of interaction and explainability in such systems. %
The lack of customisable explanations has also received attention in the literature~\cite{schneider2019personalized}. %
\citet{schneider2019personalized} have reviewed an array of explainability approaches focusing on their personalisation capabilities. %
They have observed that personalised explanations in XAI and IML are generally absent from the existing literature. %
To help researchers design and implement such methods, \citet{schneider2019personalized} proposed a generic framework for personalised explanations that identifies their three adjustable properties: complexity, content (called ``decision information'', i.e., what to explain) and presentation (how to explain, e.g., figures vs.\ text); \citet{eiband2018bringing} discussed the latter two properties from a user interface design perspective. %
Furthermore, \citet{schneider2019personalized} highlighted that interactive personalisation of explanations can either be an \emph{iterative}, e.g., a conversation, or a \emph{one\hyp{}off} process, e.g., specifying constrains before the explanation is generated. %
The latter approach does not, however, require the explainability system to be interactive as the same personalisation can be achieved off\hyp{}line by extracting the personalisation specification from the explainee and subsequently incorporating it into the data or algorithm initialisation. %
Interaction with explainability systems has also been acknowledged by \citet{henin2019towards}, who proposed a generic mathematical formulation of black\hyp{}box explainers consisting of three distinct steps: sampling, generation and \textbf{interaction}.%

While some explainability approaches introduced in the literature are simply incapable of interactive personalisation -- a number of them may still be personalised off\hyp{}line -- others are, however this property is neither utilised~\cite{wachter2017counterfactual} nor acknowledged. %
This lack of recognition may be because the explainability system designers do not see the benefits of this step or due to the difficulties with building such systems (from the engineering perspective) as well as evaluating them. %
To facilitate interactive personalisation the user interface has to be capable of delivering explanations and collecting explainees' feedback, which may require an interdisciplinary collaboration with User Experience and Human\hyp{}Computer Interaction researchers. %
Systematic evaluation and validation of this type of explainers is also more elaborate, possibly requiring multiple rounds of time\hyp{}consuming user studies.%

Despite these hurdles, a number of explainability tools and techniques enable the user to personalise explanations to some extent. %
\citet{akula2019natural} presented a dialogue\hyp{}driven explainability system that uses contrastive explanations based on predictions derived from And\hyp{}Or graphs and hand\hyp{}crafted ontology, however generalising this technique may be challenging as it requires hand\hyp{}crafting separate ontology and And\hyp{}Or graph for each application. %
\citet{lakkaraju2019faithful} introduced rule\hyp{}based explanations that the user can personalise by choosing which features will appear in the explanation -- an off\hyp{}line personalisation. %
Google published their \emph{what\hyp{}if} tool\footnote{\url{https://pair-code.github.io/what-if-tool/}}, which provides the explainee with an interactive interface that allows generating contrastive explanations of selected data points by modifying their features, i.e., asking ``What if?'' questions.%

In our work we strive to bring together the most important concepts from this wide spectrum of research as a generic and powerful aid to people building explainers of predictive systems that allow explanation personalisation via on\hyp{}line interaction. %
To this end, we provide an overview of a voice\hyp{}driven contrastive explainer built for an ML loan application model, which allows the explainee to interrogate its predictions by asking counterfactual questions~\cite{sokol2018glass}. %
We discuss our experience from building, deploying and presenting the system, which allowed us to critically evaluate its properties and formulate further desiderata and lessons learnt. %
We present these observations in the following sections as guidelines for developing similar projects.%

\section{Interactively Customisable Explanations\label{sec:chap07:glass-box}}
As a first step towards personalised, interactive XAI systems we developed \emph{Glass\hyp{}Box}~\cite{sokol2018glass}: a class\hyp{}contrastive counterfactual explainability system that can be queried with a natural language dialogue (described in Section~\ref{sec:chap07:glass-box_design}). %
It supports a range of ``Why?'' questions that can be posed either through a voice\hyp{} or chat\hyp{}based interface. %
Building this system and testing it in the wild provided us with invaluable experience and insights, which we now share with the community as they may be useful to anybody attempting to develop and deploy a similar system -- Sections~\ref{sec:chap07:glass-box_desiderata} and \ref{sec:chap07:glass-box_properties} discuss interactive explainers \emph{desiderata} and \emph{properties} respectively. %
The feedback that helped us to refine our idea of interactive XAI systems producing personalised explanations (presented in Section~\ref{sec:chap07:glass-box_feedback}) was collected while demonstrating Glass\hyp{}Box to a diverse audience consisting of both domain experts, approached during the 27\textsuperscript{th} International Joint Conference on Artificial Intelligence (IJCAI 2018), and a lay audience, approached during a local ``Research without Borders'' festival\footnote{The festival spans a wide range of research projects both in social sciences and engineering.} that is open to the public and attended by pupils from local schools. %
While at the time of presentation our system was limited to class\hyp{}contrastive counterfactual explanations personalised by (implicitly) choosing data features that the counterfactual statements were conditioned on and provided to the user in natural language, we believe that our observations remain valid beyond this particular XAI technique. %
We hope to test this assumption in our future work -- see Section~\ref{sec:chap07:discussion} for more details -- by employing the remaining four decision tree explainability modalities listed in the introduction, albeit in an XAI system refined based on our experience to date.%

\subsection{Glass\hyp{}Box Design\label{sec:chap07:glass-box_design}}%
Glass\hyp{}Box has been designed as a piece of hardware built upon the Google AIY (Artificial Intelligence Yourself) Voice Kit\footnote{\url{https://aiyprojects.withgoogle.com/voice}} -- a customisable hardware and software platform for development of voice interface\hyp{}enabled interactive agents. %
The first prototype of Glass\hyp{}Box utilised the Amazon Alexa skill Application Programming Interface, however the limitations of this platform at the time (the processing of data had to be deployed to an on\hyp{}line server and invoked via an API call) have hindered the progress and prompted us to switch to the aforementioned Google AIY Voice device. %
These recent technological advancements in automated speech\hyp{}to\hyp{}text transcription and speech synthesis provided as a service allowed us to utilise an off\hyp{}the\hyp{}shelf, voice\hyp{}enabled, virtual, digital assistant to process explainees' speech and automatically answer their questions -- something that would not have been feasible had we decided to build this component ourselves. %
We extended the voice\hyp{}driven user interface with a (textual) chat\hyp{}based web interface that displays the transcription of the conversation and its history -- to improve accessibility of the system, among other things -- in addition to allowing the explainee to type in the queries instead of saying them out loud.%

To avoid a lengthy and possibly off\hyp{}putting process of submitting (mock) personal details -- i.e., a data point -- to be predicted by the underlying Machine Learning model and explained by Glass\hyp{}Box, we opted for a predefined set of ten data points. %
Any of them could be selected and input to Glass\hyp{}Box by scanning a QR code placed on a printed card that also listed details of this fictional individual.%

\begin{figure}[htb]
    \centering
    \includegraphics[width=.45\textwidth]{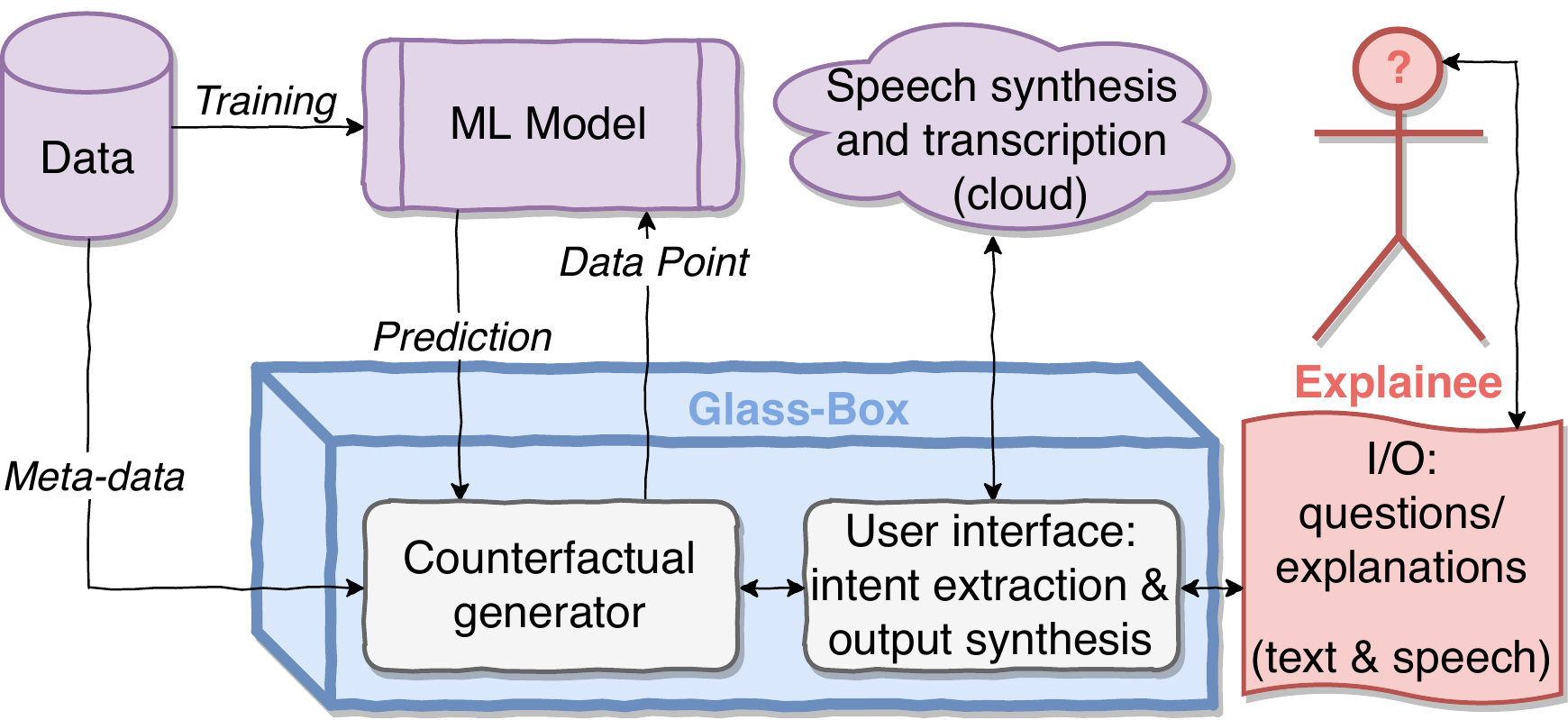}
    \caption{Glass\hyp{}Box design and information flow.}
    \label{fig:chap07:glass-box_design}
\end{figure}

Once a data point is selected, the explainee can alter personal details of this fictional individual by interacting with Glass\hyp{}Box, e.g., ``I am 27 years old, not 45.'' %
Any input to the system is passed to a natural language processing and understanding module built using \emph{rasa}\footnote{\url{https://github.com/RasaHQ/rasa}}. %
Our deployment of the Glass\hyp{}Box system was based on the \emph{UCI German credit} data set\footnote{\url{https://archive.ics.uci.edu/ml/datasets/statlog+(german+credit+data)}} (using a subset of its features) for which a decision tree classifier was trained using \emph{scikit\hyp{}learn}\footnote{\url{https://scikit-learn.org/}}~\cite{scikit-learn}. %
Since the German credit data set has a binary target variable (``good'' or ``bad'' credit score), the class contrast in the counterfactual explanations is implicit. %
Nevertheless, this could be easily generalised to a multi\hyp{}class setting by requiring the explainee to explicitly specify the contrast class, taking the second\hyp{}most likely one or providing one explanation per class. %
A conceptual design of Glass\hyp{}Box is presented in Figure~\ref{fig:chap07:glass-box_design}.%

To facilitate some of the user interactions the data set had to be manually annotated. %
This process allowed the generation of engaging natural language responses and enabled answering questions related to individual fairness. %
The latter functionality was achieved by indicating which features (and combinations thereof) should be treated as protected attributes (features), hence had a counterfactual data point conditioned on one of these features been found, Glass\hyp{}Box would indicate unfair treatment of this individual. %
This functionality could be invoked by asking ``Is the decision fair?'' question and further interrogating the resulting counterfactual explanation if one was found. %
Depending on the explainability and interactiveness requirements expected of the system, other data set annotations may be required. %
Since annotation is mostly a manual process, creating them can be time\hyp{} and resource\hyp{}consuming.%

\begin{figure}[htb]
    \centering
    \includegraphics[width=.45\textwidth]{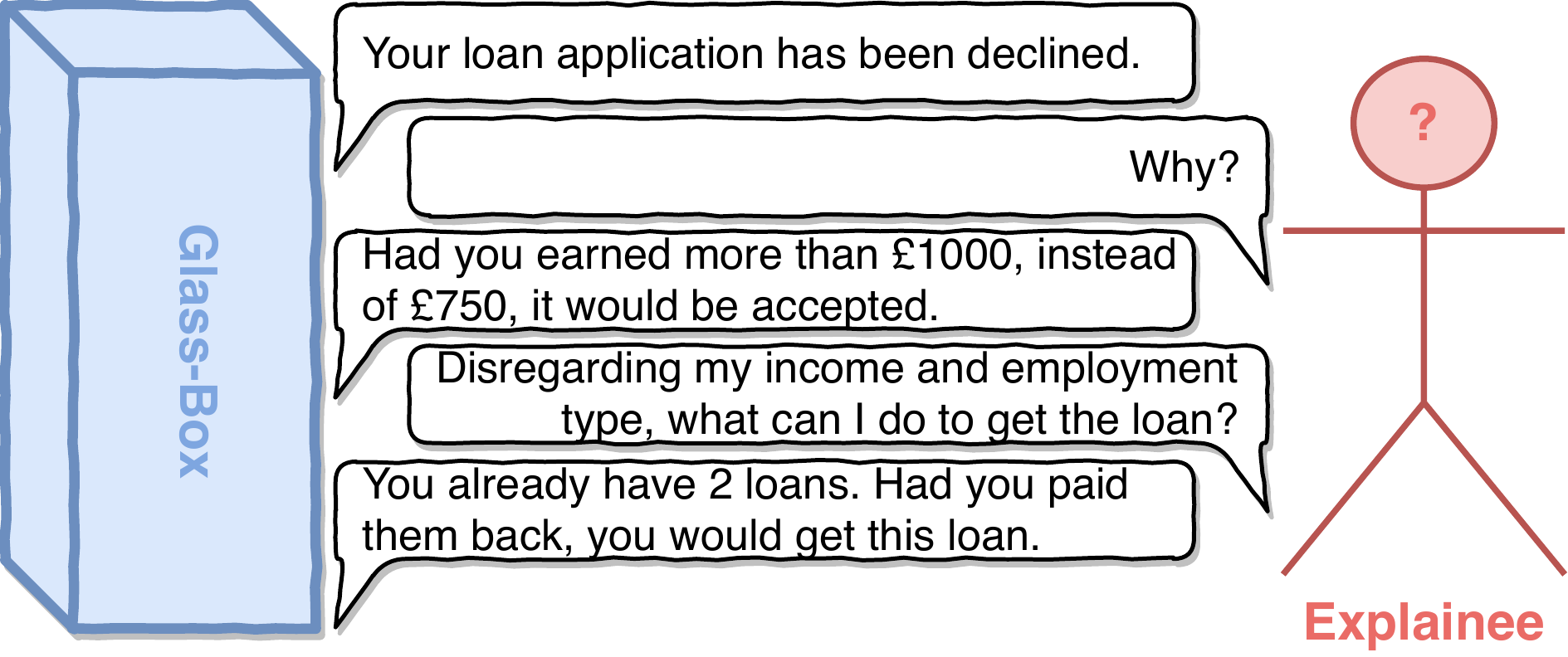}
    \caption{An example explanatory conversation between Glass\hyp{}Box and an explainee who personalises the explanations by asking counterfactual questions.}
    \label{fig:chap07:glass-box_conversation}
\end{figure}

As noted before, the main objective of Glass\hyp{}Box is to provide the users with personalised explanations whenever they decide to challenge the decision of the underlying Machine Learning model. %
The explainee can request and interactively customise the resulting counterfactual explanations through a natural language interface with appropriate dialogue cues. %
This can be done in three different ways by asking the following questions:%
\begin{itemize}%
    \item ``Why?'' -- a plain counterfactual explanation -- the system returns the shortest possible class\hyp{}contrastive counterfactual;%
    \item ``Why despite?'' -- a counterfactual explanation not conditioned on the indicated feature(s) -- the system returns a class\hyp{}contrastive counterfactual that does not use a specified (set of) feature(s) as its condition; and%
    \item ``Why given?'' -- a (partially\hyp{})fixed counterfactual explanation -- the system returns a counterfactual that is conditioned on the specified (set of) feature(s).%
\end{itemize}%
By repetitively asking any of the above ``Why?'' questions the system will enumerate all the possible explanations with the condition set (the features that need to change) increasing in quantity until no more explanations can be found. %
It is also possible to mix the latter two questions into ``Why given \ldots and despite \ldots?'', thereby introducing even stronger restrictions on the counterfactual explanations. %
In addition to ``Why?'' questions the explainee can also ask \textbf{``What if?''} %
In this case it is the user who provides the contrast and wants to learn the classification outcome of this hypothetical data point. %
This question can be either applied to the selected data point (which is currently being explained) or any of the counterfactual data points offered by the system as an explanation. %
All of these requirements imposed by the user are processed by a simple logical unit that translates the user requests into constraints applied to the set of features that the counterfactual is allowed and/or required to be conditioned upon. %
All of these happen through a natural language dialogue, an example of which is depicted in Figure~\ref{fig:chap07:glass-box_conversation}.%

The method used to generate counterfactual explanations from the underlying decision tree classifier relies upon a bespoke leaf\hyp{}to\hyp{}leaf distance metric. %
It allows to find leaves of different classes to the one assigned to the selected data point that require the fewest possible changes to this data point in its feature space. %
One obvious solution to this problem is any neighbouring leaf of a different class; this requires just one feature to be altered. %
However, there may also exist leaves that are relatively distant in the decision tree structure but also require just one feature value change, for example, when these two decision tree paths do not share many features. %
This distance metric is computed by representing the tree structure in a binary \emph{meta\hyp{}feature} space that is created by extracting all the unique feature partitions from the splits of the decision tree. %
Finally, an \(L1\)\hyp{}like metric (when a particular feature is present on one branch and absent on the other, this distance component is assumed to be \(0\)) is calculated and minimised to derive a list of counterfactual explanations ordered by their length.%

\subsection{Explanation Desiderata\label{sec:chap07:glass-box_desiderata}}
During the development stage and early trials of Glass\hyp{}Box we identified a collection of \emph{desiderata} and \emph{properties} that should be considered when building such systems. %
Some of these attributes are inspired by relevant literature~\cite{weld2018challenge,miller2018explanation,kulesza2015principles,schneider2019personalized}, while others come from our experience gained in the process of building the system, presenting it to various audiences, discussing its properties at different events and collecting feedback about interacting with it. %
While this and the following sections focus on desiderata for interactive and customisable explanations, we provide an in\hyp{}depth discussion on this topic for generic explainability systems in our work on ``Explainability Fact Sheets''~\cite{sokol2020explainability}. %
The relevant subset of these desiderata are summarised in Table~\ref{tab:chap07:glass-box_desiderata} as well as collected and discussed below. %
Section~\ref{sec:chap07:glass-box_properties}, on the other hand, examines the properties of interactive explainability systems.%

\begin{table*}[tbh]
    \centering
    \begin{tabular}{p{5.25cm}p{5.25cm}p{5.25cm}}
        \textbf{Functional} & \textbf{Operational} & \textbf{Usability} \\
        \hline
        F3: Explanation Target & O7: Function of the Explanation & U3: Contextfullness \\
        F4: Explanation Breadth/Scope & O8: Causality vs.\ Actionability & U6: Chronology \\
        F7: Relation to the Predictive System & & U7: Coherence \\
         & & U8: Novelty \\
         & & U9: Complexity \\
         & & U10: Personalisation \\
         & & U11: Parsimony
    \end{tabular}
    \caption{A subset of \emph{desiderata} for explainability systems proposed by \citet{sokol2020explainability}, which are applicable to interactive explainers that support personalisation. (Please see Section~\ref{sec:chap07:glass-box_desiderata} for a discussion.)\label{tab:chap07:glass-box_desiderata}}
\end{table*}

Given the complex nature of such systems, it would be expected that some of these objectives might be at odds with each other, their definition may be ``fuzzy'', they might be difficult to operationalise, their ``correct'' application might depend on the use case, etc. %
Furthermore, striking the right balance between these desiderata can be challenging. %
Nevertheless, we argue that considering them while designing interactive explainers will improve the overall quality of the system, help the designers and users understand their strengths and limitations, and make the interaction feel more natural to humans. %
Furthermore, some of these desired properties can be achieved (and ``optimised'' for the explainees) by simply allowing user interaction, thereby alleviating the need of explicitly building them into the system. %
For example, interactive \emph{personalisation} of the explanations (on\hyp{}line, with user input) can mean that it does not have to be solved fully algorithmically off\hyp{}line.%

The main advantage of Glass\hyp{}Box interactiveness is the explainee's ability to transfer knowledge onto the system -- in this particular case various preferences with respect to the desired explanation -- thereby \emph{personalising} the resulting explanation~\cite[property U10, see Table~\ref{tab:chap07:glass-box_desiderata}]{sokol2020explainability}. %
In our experience, personalisation can come in many different shapes and forms, some of which are discussed below. %
By interacting with the system the explainee should be able to adjust the \emph{breadth and scope} of an explanation~\cite[property F4]{sokol2020explainability}. %
Given the complexity of the underlying predictive model, the explainee may start by asking for an explanation of a \emph{single data point} (black\hyp{}box prediction) and continue the interrogation by generalising it to an explanation of a \emph{sub\hyp{}space of the data space} (a cohort explanation) with the final stage entailing the explanation of the entire black\hyp{}box model. %
Such a shift in explainee's interest may require the explainability method to adapt and respond by changing the \emph{target} of the explanation~\cite[property F3]{sokol2020explainability}. %
The user may request an explanation of a single data point or a summary of the whole data set (training, test, validation, etc.), an explanation of a predictive model (or its subspace) or any number of its predictions. %
Furthermore, interactive personalisation of an explanation can increase the overall versatility of such systems as customised explanations may serve different purposes and have different \emph{functions}~\cite[property O7]{sokol2020explainability}. %
An appropriately phrased explanation may be used as an evidence that the system is \emph{fair} -- either with respect to a group or an individual depending on the scope and breadth of the explanation -- or that it is \emph{accountable}, which again can be investigated with a varied scope, for example, a ``What if?'' question uncovering that two seemingly indistinguishable data points yield significantly different class assignment, aka adversarial examples~\cite{goodfellow2015explaining}. %
Importantly, if the explainer is flexible enough and the interaction allows such customisation, however the explanations were designed to serve only one purpose, e.g., transparency, the explainee should be explicitly warned of such limitations to avoid any unintended consequences. %
For example, the explanations may be counterfactually actionable but they are not causal as they were not derived from a causal model~\cite[property O8]{sokol2020explainability}.%

Some of the aforementioned principles can be observed in how Glass\hyp{}Box operates. %
The contrastive statements about the underlying black\hyp{}box model can be used to assess its transparency (their main purpose), fairness (disparate treatment via contrastive statements conditioned on protected attributes) and accountability (e.g., answers to ``What if?'' questions that indicate an unexpected non\hyp{}monotonic behaviour). %
The contrastive statements are personalised via user\hyp{}specified constrains of the conditional part (foil) of the counterfactual explanation and by default are with respect to a single prediction. %
Cohort\hyp{}based insights can be retrieved by asking ``What if?'' questions with regard to counterfactual explanations generated by Glass\hyp{}Box -- Section~\ref{sec:chap07:discussion} discusses how the scope and the target of our explanations can be broadened to global explanations of the black\hyp{}box model. %
Given the wide range of possible explanations and their uses some systems may produce contradictory or competing explanations. %
Glass\hyp{}Box is less prone to such issues as the employed explainer is ante\hyp{}hoc~\cite[property F7]{sokol2020explainability}, i.e., predictions and explanations are derived from the same ML model, hence they are always truthful with respect to the predictive model. %
This means that contradictory explanations are indicative of flaws in the underlying ML model, hence can be very helpful in improving its accountability.%

In day\hyp{}to\hyp{}day human interactions we are able to communicate effectively and efficiently because we share common background knowledge about the world that surrounds us -- a mental model of how to interact with the world and other people~\cite{kulesza2012tell}. %
Often, human\hyp{}machine interactions lack this implicit link making the whole process feel unnatural and frustrating. %
Therefore, the creators of interactive explainability techniques should strive to make their systems \emph{coherent} with the explainee's mental model to mitigate this phenomenon as much as possible~\cite[property U7]{sokol2020explainability}. %
While this objective may not be achievable in general, modelling a part of the user's mental model, however small, can make a significant difference. %
The two main approaches to extracting an explainee's mental model are interactive querying of the explainee in an iterative dialogue (on\hyp{}line), or embedding the user's characteristics and preferences in the data or in the parameters of the explainer (off\hyp{}line), both of which are discussed in Section~\ref{sec:chap07:related_work}.%

For explainability systems this task is possible to some extent as their operation and purpose are limited in scope in contrast to more difficult tasks like developing a generic virtual personal assistant. %
Designers of such systems should also be aware that many interactions are underlined by implicit assumptions that are embedded in the explainee's mental model and perceived as mundane, hence not voiced, for example, the context of a \emph{follow\hyp{}up} question. %
However, for human\hyp{}machine interactions the \emph{context} and its dynamic changes can be more subtle, which may cause the coherence of the internal state of an explainer and the explainee's mental model to diverge~\cite[property U3]{sokol2020explainability}. %
This issue can be partially mitigated by explicitly grounding explanations in a context at certain stages, for example, whenever the context shifts, which will help the users to adapt by updating their mental model and assumptions. %
Contextfullness will also help the explainee better understand the limitations of the system, e.g., whether an explanation produced for a single prediction can (or must not) be generalised to other (similar) instances: ``this explanation can be generalised to other data points that have all of the feature values the same but feature \(x_5\), which can span the \(0.4 \leq x_5 < 1.7\) range.''%

Regardless of the system's interactivity, the explanations should be \emph{parsimonious} -- as short as possible but not shorter than necessary -- to convey the required information without overwhelming the explainee~\cite[property U11]{sokol2020explainability}. %
Maintaining a mental model of the user can help to achieve this objective as the system can provide the explainee only with \emph{novel} explanations -- accounting for factors that the user is not familiar with -- therefore reducing the amount of information carried by the explanation~\cite[property U8]{sokol2020explainability}. %
Another two user\hyp{}centred aspects of an explanation are its \emph{complexity} and granularity~\cite[property U9]{sokol2020explainability}. %
The complexity of explanations should be adjusted according to the depth of the technical knowledge expected of the intended audience, and the level of detail chosen appropriately for their intended use. %
This can either be achieved by design (i.e., incorporated into the explainability technique), be part of the system configuration and parametrisation steps (off\hyp{}line) or adjusted interactively by the user as part of the explanatory dialogue (on\hyp{}line). %
Another aspect of an explanation, which is often expected by humans~\cite{miller2018explanation}, is the \emph{chronology} of factors presented therein: the explainee expects to hear more recent events first~\cite[property U6]{sokol2020explainability}. %
While this property is data set\hyp{}specific, the explainee should be given the opportunity to trace the explanation back in time, which can easily be achieved via interaction.%

Glass\hyp{}Box attempts to approximate its users' mental models by mapping their interests and interaction context (inferred from posed questions) to data features that are used to compose counterfactual explanations. %
Memorising previous interactions, their sequence and the frequency of features mentioned by the user help to achieve this goal and avoid repeating the same answers -- once all of the explanations satisfying given constraints were presented, the system explicitly states this fact. %
Contextfullness of explanations is based on user interactions and is implicitly preserved for follow\hyp{}up queries in case of actions that do not alter the context and are initiated by the user -- e.g., interrogative dialogue. %
Whenever the context shifts -- e.g., a new personalised explanation is requested by the user or an interaction is initiated by Glass\hyp{}Box -- it is explicitly communicated to the user. %
Contrastive explanations are inherently succinct, but a lack of parsimony could be observed for some of Glass\hyp{}Box explanations, which resulted in a long ``monologue'' delivered by the system. %
In most of the cases this was caused by the system ``deciding'' to repeat the personalisation conditions provided by the user to ensure their coherence with the explainee's mental model.%

Glass\hyp{}Box is capable of producing novel explanations by using features that have not been acknowledged by the user during the interaction. %
Interestingly, there is a trade\hyp{}off between novelty of explanations and their coherence with the user's mental model, which we have not explored when presenting our system but which should be navigated carefully to avoid jeopardising explainee's trust. %
Glass\hyp{}Box was built to explain predictions of the underlying ML model and did not account for possible generalisation of its explanations to other data points (the users were informed about it prior to interacting with the device). %
However, the explainees can ask ``What if?'' questions with respect to the counterfactual explanations, e.g., using slight variations of the explained data point, to explicitly check whether their intuition about the broader scope of an explanation holds up. %
Finally, chronology was not required of Glass\hyp{}Box explanations as the data set used to train the underlying predictive model does not have any time\hyp{}annotated features.%

\subsection{Glass\hyp{}Box Properties\label{sec:chap07:glass-box_properties}}
In addition to a set of interactive explainability system desiderata, we consider a number of their general properties and requirements that should be considered prior to their development. %
These are summarised in Table~\ref{tab:chap07:glass-box_properties} and discussed below.%

\begin{table*}[tbh]
    \centering
    \begin{tabular}{p{5.25cm}p{5.25cm}p{5.25cm}}
        \textbf{Operational} & \textbf{Usability} & \textbf{Safety} \\
        \hline
        O1: Explanation Family & U4: Interactiveness & S3: Explanation Invariance \\
        O2: Explanatory Medium & & \\
        O3: System Interaction & & \\
        O4: Explanation Domain & & \\
        O5: Data and Model Transparency & & \\
        O6: Explanation Audience & & \\
        O10: Provenance & &
    \end{tabular}
    \caption{A subset of \emph{properties} of explainability systems proposed by \citet{sokol2020explainability}, which are applicable to interactive explainers that support personalisation. (Please see Section~\ref{sec:chap07:glass-box_properties} for a discussion.)\label{tab:chap07:glass-box_properties}}
\end{table*}

Assuming that the system is interactive, the \emph{communication protocol} between the explainee and the explainer should be carefully chosen to support the expected input and deliver the explanations in the most natural way possible. %
For example, clearly indicating which parts of the explanation can be personalised and the limitations of this process should be disclosed to the user~\cite[property O3, see Table~\ref{tab:chap07:glass-box_properties}]{sokol2020explainability}. %
The choice of \emph{explanatory medium} used to convey the explanation is also crucial. %
Plots, interactive or not, can be very informative, but may not convey the whole story due to the curse of dimensionality and the limitations of the human visual system~\cite[property O2]{sokol2020explainability}. %
Supporting visualisations with textual description can greatly improve their intelligibility, and vice versa, nevertheless in some cases this approach may be sub\hyp{}optimal, for example, explaining images using only a natural language interface. %
The \emph{intended audience} should be considered in conjunction with the communication protocol to choose a suitable explanation type~\cite[property O6]{sokol2020explainability}. %
Domain experts may prefer explanations expressed in terms of the internal parameters of the underlying predictive model, but a lay audience may rather prefer exemplar explanations that use relevant data points -- choosing the appropriate \emph{explanation domain}~\cite[property O4]{sokol2020explainability}. %
The audience also determines the purpose of the explanation. %
For example, inspecting a predictive model for debugging purposes will need a different system than guiding the explainee with an actionable advice towards a certain goal like getting a loan. %
Interactive explainers can support a wide spectrum of these properties by allowing the explainee to personalise the output of the explainer as discussed in Section~\ref{sec:chap07:glass-box_desiderata}.%

Achieving some of these objectives may require the features of the underlying data set or the predictive model itself to be \emph{transparent}~\cite[property O5]{sokol2020explainability}. %
For example, consider explaining a model trained on a data set with features that are object measurements in meters in contrast to magnitudes of embedding vectors. %
When the raw features (original domain) are not human\hyp{}interpretable, the system designer may decide to use an interpretable representation (transformed domain) to aid the explainee. %
Providing the users with the \emph{provenance} of an explanation may help them to better understanding its origin, e.g., an explanation purely based on data, model parameters or both~\cite[property O10]{sokol2020explainability}. %
Choosing the right \emph{explanation family} is also important, for example: relation between data features and the prediction, relevant examples such as similar data points or causal mechanisms~\cite[property O1]{sokol2020explainability}. %
Again, interactive explainers have the advantage of giving the user the opportunity to switch between multiple different explanation types. %
Furthermore, the design of the user interface should be grounded in the Interactive Machine Learning, Human\hyp{}Computer Interaction, User Experience and Explainable Artificial Intelligence research to seemingly deliver the explanations. %
For example, the explainee should be given the opportunity to reverse the effect of any actions that may influence the internal state of the explainer and the system should always respect user's preferences and feedback~\cite[property U4]{sokol2020explainability}. %
Finally, if an explanation of the same event can change over time or is influenced by a random factor, user's trust is at stake. %
The explainee should always be informed about the degree of \emph{explanation invariance} and its manifestation in the output of an explainer~\cite[property S3]{sokol2020explainability}. %
This property is vital to Glass\hyp{}Box's success, which we discuss in more detail in Section~\ref{sec:chap07:discussion}.%

\subsection{Glass\hyp{}Box Reception and Feedback\label{sec:chap07:glass-box_feedback}}
We presented Glass\hyp{}Box to domain experts (general AI background knowledge) and a lay audience with the intention to gauge their reception of our prototype and collect feedback that would help us revise and improve our explainability system. %
To this end, we opted for informal and unstructured free\hyp{}form feedback, which was mostly user\hyp{}driven and guided by reference questions (based on our list of desiderata) whenever necessary. %
We decided to take this approach given the nature of the events at which we presented our prototype -- a scientific conference and a research festival.%

Glass\hyp{}Box is composed of multiple independent components, all of which play a role in the user's reception of the system:%
\begin{itemize}%
    \item natural language understanding and generation,%
    \item speech transcription and synthesis,%
    \item voice and text user interfaces, and%
    \item a data set that determines the problem domain.%
\end{itemize}%
Therefore, collecting free\hyp{}form feedback at this early stage helped us to pin\hyp{}point components of the system that required more attention and identify possible avenues for formal testing and design of user studies.%

While presenting the device we only approached members of the audience who expressed an interest in interacting with the device and who afterwards were willing to describe their experience. %
In total, we collected feedback from 6 domain experts and 11 participants of the research festival of varying demographics. %
When introducing the system and its modes of operation to the participants, we assessed their level of AI and ML expertise by asking background questions, which allowed us to appropriately structure the feedback session.%

While discussing the system with the participants, we were mainly interested in their perception of its individual components and suggestions about how these can be improved. %
Most of the participants enjoyed asking questions and interacting with the device via the voice interface, however some of them found the speech synthesis module that answered their questions ``slow'', ``unnatural'' and ``clunky''. %
These observations have prompted some of the participants to disable voice\hyp{}based responses and use the text\hyp{}based chat interface to read the answers instead of listening to them. %
When asked about the quality of explanations, their comprehensibility and content, many participants were satisfied with received answers. %
They claimed that personalised explanations provided them with information that they were seeking for as opposed to the default explanation given at first. %
However, some of them expressed concerns regarding the deployment of such systems in everyday life and taking the human out of the loop. %
The most common worry was the impossibility to ``argue'' and ``convince'' the explainer that the decision is incorrect and the explanation does not capture the complexity of one's case. %
Some participants were also sceptical of the general idea of interacting with an AI agent and the fail\hyp{}safe mode of the device, which produced ``I cannot help you with this query.'' response whenever the explainer could not answer the user's question.%

We plan to use all of this feedback and our experience in building interactive explainers to refine the system focusing on its explanation personalisation aspect, and test this particular component with formal user studies. %
Isolating this module of the explainer will alleviate the influence of the user interface on the perception of the explanations, allowing us to investigate the effectiveness and reception of personalised explanations in a formal setting.%

\section{Discussion\label{sec:chap07:discussion}}%
Developing Glass\hyp{}Box and demonstrating it to a diverse audience provided us with a unique experience of building, deploying and refining interactive explainers. %
To help researchers and engineers with a similar agenda we summarise the lessons learnt in Section~\ref{sec:chap07:lessons}. %
We also discuss our next steps in interactive and personalised explainability research in Section~\ref{sec:chap07:future} to draw attention to interesting open questions.%

\subsection{Lessons Learnt\label{sec:chap07:lessons}}
The major challenge of building Glass\hyp{}Box was the development overhead associated with setting up the hardware and software needed to make it voice\hyp{}enabled and capable of processing the natural language. %
While ready\hyp{}made components were adapted for these purposes, the effort required to build such a system is still significant which may not always be justified. %
We encourage researchers to build such a system if the research value lies in the system itself or it is used as a means to an end, for example, research on interactive explainability systems. %
In this case, one should be aware of generalisability issues as each new data set used within such a system must be adapted by preparing appropriate annotations and (possibly) training a new natural language processing model. %
In many cases, based on our observations, it seems that all this effort is only justified when the creator of the system is committed to deploying it in real life. %
For research purposes, however, the engineering overhead can be overwhelming, in which case we suggest using the \emph{Wizard of Oz} studies~\cite{dahlback1993wizard} as an accessible alternative.%

Once Glass\hyp{}Box was operational, its major usability barrier was the time\hyp{}consuming process of inputting personal data when role\hyp{}playing the loan application process. %
At first, we implemented this step as a voice\hyp{}driven question\hyp{}answering task but even with just 13 attributes (most of which were categorical) this proved to be a challenge for the explainee. %
We overcame this issue by pre\hyp{}defining 10 individuals whom the explainee could impersonate. %
We then allowed the explainee to further customise the attributes of the selected individual by asking Glass\hyp{}Box to edit them (with voice\hyp{} and text\hyp{}based commands). %
In hindsight, we believe that this kind of task should be completed by using a dedicated input form (e.g., a questionnaire delivered as a web page), thereby giving the explainee the full control of the data input process and mitigating the lengthy ``interrogation'' process.%

The interactive aspect of Glass\hyp{}Box (discussed in length in Section~\ref{sec:chap07:glass-box_design}) provides many advantages from the point of view of explainability. %
For example, it enables the explainee to assess individual fairness of the underlying predictive model and personalise the explanations (see Sections~\ref{sec:chap07:glass-box_desiderata} and~\ref{sec:chap07:glass-box_properties} for more details). %
However, not all types of explainability algorithms allow for the resulting explanation to be interactively customised and personalised, restricting the set of tools that can be deployed in such a setting. %
If incorporating the user feedback (delivered as part of the interaction, e.g., via argumentation~\cite{madumal2019grounded}) into the underlying predictive model is desired, this model has to support refinements beyond the training phase, further reducing the number of applicable Machine Learning and explainability techniques.%

As noted in Section~\ref{sec:chap07:glass-box_desiderata}, some of the interactivity and personalisation desiderata cannot be achieved without ``simulating'' the explainee's mental model. %
While we believe that cracking this problem will be a corner stone of delivering explanations that feel natural to humans, we do not expect it to be solved across the board in the near future.%

In case of Glass\hyp{}Box, where the explanations are presented to the user as counterfactual statements, we observed a tendency amongst the explainees to generalise an explanation of a single data point to other, relatively similar, instances. %
However tempting, Glass\hyp{}Box explanations cannot be generalised as they are derived from a predictive model (structure of a decision tree) that does not encode and account for the causal structure of the underlying phenomenon. %
This can sometimes lead to contradictory explanations, which can be detrimental to the explainee's trust. %
Since Glass\hyp{}Box uses an ante\hyp{}hoc explainability algorithm (i.e., explanations and predictions are derived from the same ML model), contradictory, incorrect or incoherent explanations are indicative of issues embedded in the underlying predictive model, which should be reported to and addressed by the model creators. %
However, if a post\hyp{}hoc explainability tool is employed (explanations are not derived directly from the predictive model, e.g., surrogate explainers), contradictory explanations manifest a problem with the system. %
This issue cannot be uniquely pinpointed and can either be attributed to a low\hyp{}fidelity explainer or to an underperforming predictive model, putting explainees' trust at risk. %
Communicating the limitations of the explanations clearly can help to partially mitigate this problem; grounding the explanations in a context (see Section~\ref{sec:chap07:glass-box_desiderata}) is another approach.%

While truthful to the underlying black box, an ante\hyp{}hoc explainability approach may not be available for a chosen predictive model. %
For example, deep neural networks are intrinsically complex, which encumbers explaining them without resorting to proxies. %
This observation highlights another trade\hyp{}off of AI\hyp{}based systems: predictive power vs.\ transparency~\cite{gunning2019darpa}. %
Simpler models such as decision trees are less expressive but more interpretable. %
On the other hand, complex models such as deep neural networks are more powerful at the expense of opacity. %
It is still possible to explain the latter model family with proxies and post\hyp{}hoc approaches, but issues with the fidelity and truthfulness of such explanations can be unacceptable, e.g., in high\hyp{}stakes situations such as criminal justice or financial matters~\cite{rudin2019stop}. %
These conclusions have led some researchers (e.g., \citet{rudin2019stop}) to deem post\hyp{}hoc explainers as outright harmful. %
Instead, they argue, developers behind predictive systems for high\hyp{}stake applications should invest more time in feature engineering and restrict their toolkit to inherently transparent ML models.%

As might be expected, the power and flexibility of Glass\hyp{}Box explanations comes at a cost. %
The interactiveness of the process enables malicious users to ask for explanations of arbitrary data points, which in large quantities may expose internals of the underlying predictive model~\cite[properties S1 and S2]{sokol2020explainability}. %
Adversaries can misuse the information leaked by the system in an attempt to reverse\hyp{}engineer the underlying predictive model (which may be proprietary) or use this knowledge to game it. %
This is particularly visible in Glass\hyp{}Box as the condition of the counterfactual explanations is based on one of the splits in the underlying decision tree, thereby revealing the exact threshold applied to one of the features, e.g., ``had you been older than 25,\dots'' implies \texttt{age > 25} internal splitting node. %
Since every explanation reveals a part of the tree structure (at least one split), with a certain budget of queries the adversary can reconstruct the tree.%

This issue is intrinsic to ante\hyp{}hoc explainers but may also affect high\hyp{}fidelity post\hyp{}hoc approaches, albeit to a lesser extent since in the latter case the explanations are not generated directly from the black\hyp{}box model. %
This undesired side\hyp{}effect can be controlled to some extent by limiting the explanation query budget for untrustworthy users or obfuscating the precise (numerical) thresholds. %
The latter can be achieved either by injecting random noise (possibly at the expense of explainees' trust) or replacing the numerical values with \emph{quantitative adjectives}, e.g., ``slightly older'' (shown to enhance user satisfaction~\cite{biran2014justification}). %
The trade\hyp{}off between transparency and security of interactive explainers should be explicitly considered during their design stage, with appropriate mitigation technique implemented and documented.%

\subsection{Next Steps\label{sec:chap07:future}}
One of the main contributions of Glass\hyp{}Box lies in the composition of its software stack and hardware architecture. %
While investigating the challenge of readying such a system for a deployment is one possible avenue for future research, we believe that a more interesting direction is to design explainability tools and techniques that facilitate (interactive) personalisation of their explanations. %
Since the latter research aspect is conditioned upon the availability of the former, we plan to use the Wizard of Oz approach~\cite{dahlback1993wizard} to mitigate the need for building an interactive user interface that is responsible for processing the natural language. %
In this scenario, the input handling and the output generation are done by a human disguised as an intelligent interface, who can access all the components of the tested explainability approach and is only allowed to take predefined actions. %
Therefore, bypassing an algorithmic natural language interface by using the Wizard of Oz~\cite{dahlback1993wizard} approach will allow us to focus our research agenda on designing and evaluating the properties of personalised explanations. %
It will also ensure that our findings are not adversely affected by poor performance of the natural language interface.%

To this end, we will use a bespoke surrogate explainer of black\hyp{}box predictions similar to the Local Interpretable Model\hyp{}Agnostic Explanations (LIME)~\cite{ribeiro2016why} algorithm. %
Our explainer uses decision trees as the local surrogate model~\cite{sokol2019blimey}, whereas LIME is based on a sparse linear regression. %
Among others, this modification allows us to improve the fidelity of the explainer by reducing the number of conflicting explanations. %
Furthermore, a tree\hyp{}based surrogate inherits the best of both worlds: the explainer is model\hyp{}agnostic, hence it can be used with any black\hyp{}box model, and it can take advantage of the wide variety of explanations supported by decision trees (as discussed in Section~\ref{sec:chap07:intro}).%

We plan to apply this approach to three different data domains: tabular data, text, and images, which tests its capacity of interactively generating personalised explanations for a range of tasks. %
We expect object recognition for images and sentiment analysis for text to be the most effective evaluation tasks as they do not require any background knowledge. %
In our studies, the explainees will be asked to interactively personalise two aspects of the explanations: an interpretable representation of the data features and their content.%

The objective of the first task depends on the data domain. %
For text, it will allow the explainees to adjust and introduce the concepts that can only be expressed with multiple words since the default interpretable representation would be a bag of words. %
For images, the users will modify the default super\hyp{}pixel segmentation to separate semantically meaningful regions -- see the example shown in Figure~\ref{fig:chap07:segmentation} and discussed in Section~\ref{sec:chap07:intro}. %
For tabular data, the interpretable representation is achieved by discretising continuous features. %
Since the local surrogate model is a decision tree, this representation is learnt automatically and cannot be explicitly modified by the explainee. %
We will give the users indirect control over the feature splits by allowing them to adjust the tree structure in terms of depth, the number of data points required for a split and the minimum number of data points per leaf.%

The second personalisation objective will allow the explainee to choose the explanation type and customise it accordingly. %
The visualisation of the surrogate tree structure can either depict the whole tree or zoom in on its selected part. %
The explainee will also be able to inspect tree\hyp{}based feature importance either by viewing all of them in a list or by querying the importance of the chosen ones. %
These two explanation types will allow the user to grasp the overall behaviour of the black\hyp{}box model in the vicinity of the explained data point. %
For text and images these will be the interactions between the words and super\hyp{}pixels in that region, i.e., within a sentence and an image respectively, and for tabular data the influence of raw features and ranges of their values. %
Furthermore, the explainee will be able to get personalised explanations of individual predictions. %
A counterfactual retrieved from the local tree, e.g., ``had these two super\hyp{}pixels/words not been there, the image/sentence would be classified differently'', will be customised by specifying constraints appearing in its condition. %
Next, the explainee can request a logical rule, e.g., ``these three super\hyp{}pixels/words must be present and these two must be removed'', for any leaf in the tree, which will be extracted from the corresponding root\hyp{}to\hyp{}leaf path. %
Both of these explanations will allow the user to understand how parts of an image or a sentence (super\hyp{}pixels and words respectively) come together to predict a data point. %
Finally, the user can view exemplar explanations of any prediction, which come from the part of the surrogate training set -- generated by perturbing the selected data point, possibly in the interpretable domain -- assigned to the relevant tree leaf. %
This means that the exemplars will be images with occluded super\hyp{}pixels, sentences with missing words and, for tabular data, slight variations of the explained data point in its original feature space. %
We believe that this diverse set of personalised explanations will encourage the user to investigate different aspects of the black\hyp{}box model thus lead to a better understanding of its behaviour.%

Interactions form another aspect of a system that delivers a multitude of different explanation types. %
A user who has learnt which features are important may want to know whether one of the counterfactual explanations is conditioned upon them. %
In particular, we want to investigate whether the user would discount the counterfactual explanations conditioned upon unimportant features and focus on the ones that include important factors. %
Also, we are interested in how the user's confidence is affected upon discovering that most of the (counterfactual) explanations are based on features indicated as unimportant by a different explanation type.%

As part of the study we aim to recognise current limitations of the interaction and personalisation aspects of the system by taking note of the requests that failed from the user's perspective. %
The possibility of retrieving multiple counterfactuals of the same length (the same number of conditions) brings up the question of their ordering. %
One approach is to use a predefined, feature\hyp{}specific ``cost'' of including a condition on that feature into the explanation. %
This heuristic can be based on the purity (accuracy) of the counterfactual leaf, the cumulative importance of features that appear on the corresponding root\hyp{}to\hyp{}leaf path, the collective importance of features listed in its conditional statement or, simply, the number of training data points falling into that leaf~\cite{waa2018contrastive}. %
However, a more user\hyp{}centred approach is to allow the explainee to supply this information either implicitly or explicitly during the interaction.%

To improve the quality of the interactions, we will build a partial mental model of the explainee using a formal argumentative~\cite{dung2009assumption} dialogue introduced by \citet{madumal2019grounded}. %
Many user arguments can be parsed into logical requirements, allowing for further personalisation and more convincing explanations. %
The roles in this dialogue can also be reversed to assess and validate the explainee's understanding of the black\hyp{}box model -- the machine questioning the human~\cite{walton2011dialogue,walton2016dialogue}. %
In this interrogative dialogue, if an insight about the black box voiced by the user is incorrect, the system can provide a personalised explanation in an attempt to correct explainee's beliefs. %
Asking the user ``What if?'' questions can further assist in this task by directing the explainee's attention towards evidence relevant to the preconceived misconceptions. %
When an interaction is finished, a succinct excerpt summarising the whole explanatory process (similar to a court transcript) can be provided to the user as a reference material. %
This document should only contain explanations that the user has challenged or investigated in detail, avoiding the ones that agree with the explainee's mental model.%

The mental model can also be utilised to adjust the granularity and complexity of explanations. %
For example, a disease can be explained in medical terms -- e.g., on a bacterial level -- or with easily observable external symptoms -- e.g., cough and abnormal body temperature -- depending on the audience. %
While solving this task for a generic case is currently not feasible, we will investigate possible approaches for a data set that exhibits a hierarchy of low\hyp{}level features, which can be hand\hyp{}crafted and incorporated into the explainer.%

\section{Summary and Conclusions}
In this paper we discussed the benefits that Interactive Machine Learning can bring to eXplainable Artificial Intelligence and Interpretable Machine Learning. %
We showed how personalised explanations can improve the transparency of a Machine Learning model and how they can be generated via a human\hyp{}machine interaction. %
While other aspects of an explainability system can also be made interactive, we argued that one of the major benefits comes from personalisation. %
In particular, we showed the difference between interactiveness of an explainability system -- e.g., interactive plots -- and interactiveness of an explanation -- e.g., personalisation. %
We supported our discussion and claims with experience gained from building and demonstrating Glass\hyp{}Box: a class\hyp{}contrastive counterfactual explainability system that communicates with the user via a natural language dialogue. %
To the best of our knowledge, this is the first XAI system tested in the wild that supports explanation customisation and personalisation via interaction.%

Our experience building Glass\hyp{}Box and experimenting with it helped to identify a collection of desired functionality and a set of properties that such systems should have. % 
We discussed which ones are applicable to Glass\hyp{}Box, and summarised a list of lessons that we have learnt. %
The most important one draws attention to the engineering overhead required to build such a system despite adapting many off\hyp{}the\hyp{}shelf components. %
We concluded that one should avoid this effort in favour of Wizard of Oz studies when the main objective is to use such a platform as a test bed for various explainability techniques, unless the intention is to deploy it afterwards. %
Other key observations concerned both the importance and impossibility of simulating an explainee's mental model. %
While doing so is highly beneficial, fully satisfying this requirement is out of reach at present. %
Nevertheless, we observed that by using a formal argumentation framework to model a part of the user\hyp{}machine interactions, it may be possible to extract some relevant knowledge from the explainee, which can be utilised to this end.%

To ground our study we have reviewed relevant literature, where we identified three related research strands and showed how our work has the potential to bridge them together. %
Our investigation has shown that while some of the explainability algorithms and tools are capable of explanation personalisation via user interaction, many more are not. %
A number of other explainability approaches, such as contrastive explanations, can easily support such interactions, however their implementations lack this functionality. %
Finally, these observations combined with our findings helped us to devise next steps for our research, which pivot around investigating personalised explanations and their properties in a more principled way.%

% BibTeX users please use one of
%\bibliographystyle{spbasic}      % basic style, author-year citations
%\bibliographystyle{spmpsci}      % mathematics and physical sciences
%\bibliographystyle{spphys}       % APS-like style for physics
%%%%%%%%%%%%%%
\balance
\bibliography{template.bib}   % name your BibTeX data base

\end{document}